# Robotic Exercise Trainer:
# How Failures and T-HRI Levels Affect User Acceptance and Trust[*]

M. Krakovski[1], N. Aharony[1], and Y. Edan[1]

*Abstract*—Physical activity is important for health and wellbeing, but only few fulfill the World Health Organization's criteria for physical activity. The development of a robotic exercise trainer can assist in increasing training accessibility and motivation. The acceptance and trust of users are crucial for the successful implementation of such an assistive robot. This can be affected by the transparency of the robotic system and the robot's performance, specifically, its failures. The study presents an initial investigation into the transparency levels as related to the task, human, robot, and interaction (T-HRI), with robot behavior adjusted accordingly. A failure in robot performance during part of the experiments allowed to analyze the effect of the T-HRI levels as related to failures. Participants who experienced failure in the robot's performance demonstrated a lower level of acceptance and trust than those who did not experience this failure. In addition, there were differences in acceptance measures between T-HRI levels and participant groups, suggesting several directions for future research.

## I. INTRODUCTION

Physical inactivity or insufficient physical activity is defined as a leading risk factor for mortality and non-communicable diseases [1], [2]. Despite the well-established benefits of physical activity, only 23% of the worldwide adult population meet the World Health Organization's criteria for physical activity [2]. This is a result of lack of motivation, resources, and access. In light of the global shortage of caregivers, physiotherapists, and trainers [3], social assistive robots have become an intriguing and potentially useful solution to fill the existing gap. Robotic exercise trainers do not have constraints of time schedule and location as human trainers do. Furthermore, they also provide a solution for the social distancing requirements.

Studies have shown that robotic trainers can motivate users to perform physical training [4]. The success of social assistive robots, including robotic trainers, depends on users' acceptance and trust [4]. A user who does not trust or accept the robot will simply not use it and miss out on the potential benefits of it. The field of human-robot interaction (HRI) is gaining increasing attention in recent years as researchers seek to improve the design of interactions for greater acceptance [5]. There are several factors that can affect HRI; in this study we analyze the effect of transparency and how failures influence this effect in a robotic trainer scenario.

Transparency is an important component of a trustworthy system [6]. Incorporating transparency into robotic designs can improve HRI as it provides shared awareness between users and robotic systems [7]. Robots can present users with differing amounts and rates of information, which is referred to as levels of transparency (LOT) [8]. The information that is provided by the robot can be classified by environment, task, human, robot and interaction [8]. LOT preferences can vary between different types of environments, tasks and robots [9], younger and older users [6], [8], [10] as well as between interaction types [11]. In this study we focused on three LOT (low, medium, and high) with information of the task as related to the human, robot, and interaction. In each different LOT, the robot adjusts the interaction and the task (i.e., the specific training exercise) according to human information and performance. These interaction changes are referred to as T-HRI levels (task-human-robot-interaction); the levels include *task adjustment according to human interaction and performance* and *define different information that the robot provides*. Previous work in transparency [11] focused only on information design without adjustments of the robot interaction. The research showed a link between the user preferred amount of information and their trust in the robot.

In the case of robot failures, transparency is particularly important in order to increase trust [6], [7]. Considering that humans make mistakes constantly, it is only logical that robots developed and operated by humans would also make errors. Robots can fail even when well designed and extensively tested. Design and programming mistakes, users who do not understand how to operate the robot, or those who use it in an improper manner, all can result in failures [12]. Studies showed that failures in robot performances affects the subjective evaluation of the HRI [13], [14]. Therefore, it is critical to consider how robots' error affect users in the design of robots. Such HRI design can help users overcome the inevitable failures so users will be willing to use these robots in the future even after they encounter a failure. This is especially important when it comes to social assistive robots (e.g. [15]). A previous video study that focused on evaluation of a user interacting with a robot [6], revealed that when the robot did not make errors, higher transparency led to higher trust when using the robot. But when the robot made mistakes, the trust was low even in the high transparency level. However, since a robot is a physical entity it is important to evaluate robots in actual use [5], interacting with a moving robot is often different from watching a video or interacting with a simulator [16], [17].

This study aims to examine if and how different T-HRI levels affect the acceptance of a robotic trainer in systems with and without failures. The Technology Acceptance Model (TAM) [18] was used to analyze acceptance. The purpose of this preliminary study was to ensure a robust experimental design before conducting experiments with elderly participants. The participants in this study were young adult

[*]This research was supported by Ben-Gurion University of the Negev through the Agricultural, Biological and Cognitive Robotics Initiative, the Marcus Endowment Fund, and the W. Gunther Plaut Chair in Manufacturing Engineering.

[1]Industrial Engineering & Management Department. Ben-Gurion University of the Negev, Beer Sheva, Israel.

students who have easier access to participate in our university experiments.

## II. METHODS

### A. Robotic Exercise Trainer

We employed "Gymmy" [19], a robotic trainer for upper body exercises developed to motivate older adults to engage in physical activity. While Gymmy demonstrates the exercises, the user's performances are monitored by a RGB-D camera with skeleton tracking software (Figure 1). Based on the software's detection of the user's performance, the robot provides feedback during training. A training session with the robot includes several different exercises. Before each exercise is performed, the robot provides verbal instructions as well as the number of repetitions that need to be completed. The robot demonstrates and performs the exercise as many times as the user is required to do so. The robot counts each correct repetition the user completes during this time.

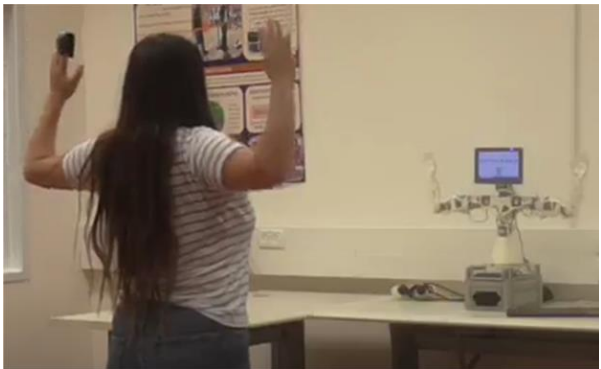

Figure 1 – Photo of a participant exercise with Gymmy

In this study the robot was enhanced with three different T-HRI levels: we developed different robot's behaviors that were matched to different transparency levels, with information of the task, human, robot, and interaction. The information in each level builds upon that in the previous level, and as the levels increase, the physical training program becomes more transparent and adapted to the user. The levels were implemented in the robotic trainer as follows:

- Low level: Basic information of the essential parts of the task and the actions of the robot as was developed initially – the robot demonstrates the exercise and provides instructions. The number of the repetitions for each exercise is chosen randomly, independent of the robot's knowledge of the human. An example of the robot instructions: "Raise your hands horizontally 8 times".

- Medium level: Adding robot information regarding its reasoning to choose the exercises plan. This plan depends on the information provided by the user to the robot at the beginning of the training. The number of repetitions for the exercises and the use of weights are determined according to this information. The robot's instructions are changed from the perviuos level, for example: "Since you are training frequently, each exercise will be performed 10 times".

- High level: Additional information is added related to the interaction and the predicted outcomes. In this level, the robots use the knowledge of the human performance during the exercises. The robot starts with a single demonstration of all the exercises, only then the training session begins. Between the exercises the number of repetitions changes based on the user's performances. The robot notes the user the reason for the change: "Due to your good performance, we will repeat the next exercise 12 times".

The original training program, intended for older adults, was adjusted for younger adults by introducing more challenging exercises, increasing repetitions of each exercise, and adding the option of performing the exercises with weights.

### B. Experimental Design

Each participant experiences three training sessions, one for each T-HRI level. The sessions were introduced to the user in a different order ensuring counterbalancing (prior to the experiment, the participants were divided into six groups, each assigned a different order). The experiment was designed as a within subject design with technology acceptance measures as dependent variables and T-HRI levels as independent variables.

### C. Experimental Procedure

After each participant filled a consent form and pre-trial questionnaires, they were presented with the robot. Each of the three sessions began by standing in front of the robot. The first session (regardless of the T-HRI level) started with an introduction of the robot and background questions, that are answered by the user using the robot's touch screen. After these steps the two remaining training sessions began. During the physical training, the robot demonstrates and explains the exercises, according to the T-HRI level. A session included three exercises. The participants filled out a post-trial questionnaire after each session, and a final questionnaire after all three sessions.

### D. Participants

51 students (16 males and 35 females) aged 23-28 ($\mu=25.67$, $\sigma=1.28$) participated in the experiments. Some of the participants encountered technical difficulties during training sessions. These problems were not controlled and occurred since their performance was not correctly recognized by the robot (due to height differences/position or location in the room). These participants expressed their frustration in the final questionnaire by mentioning that the robot should improve its user movement recognition. Overall, 24 participants (19 females and 5 males) indicated they had experienced problems with movement recognition – this group is referred as to "Failed Robot" (aged 24-28, $\mu=25.54$, $\sigma=1.22$). The group of the other 27 participants (16 females, 11 males) who experienced successful robot training is referred to as the "Successful Robot" group (aged 23-28, $\mu=25.78$, $\sigma=1.34$).

### E. Measures and Analysis

The Technology Acceptance Model (TAM) [18], was used to analyze acceptance using post-trial questionnaires similar to [16], [19]. The measures were perceived usefulness, ease-of-use and attitude. Ease-of-use was calculated as the average of comfortability and understanding. Attitude was calculated as the average of engagement, trust, satisfaction, and enjoyment.

For each measure, average and standard deviation were calculated by T-HRI level and robot performance groups ("Failed Robot" and "Successful Robot"). Statistical tests were performed in Python: Mann-Whitney Tests to compare between groups in each T-HRI level and Friedman Tests to compare between T-HRI levels.

## III. RESULTS

### A. Perceived Usefulness

For all three T-HRI levels the perceived usefulness among the participants that experienced a failed robot was lower than the ones that experienced with a successful robot (**Figure 2**). In the "Failed Robot" group the perceived usefulness medians were 3 in all levels compared to the medians in the "Successful Robot" group that were 3.5 in all three levels. However, these differences were not significant in all three levels ($p_{LowLevel}=0.1$, $p_{MediumLevel}=0.15$, $p_{HighLevel}=0.03$). In both groups the perceived usefulness was ranked highest in the medium T-HRI level, but in the "Failed Robot" group at the high T-HRI level the perceived usefulness was ranked lowest.

### B. Ease-of-use

In the high T-HRI level the median of the ease-of-use in both groups was 4.17 (**Figure 2**). In the low and medium levels, the ease-of-use medians were slightly lower in the "Failed Robot" group ($Mdn_{LowLevel}=4.17$, $Mdn_{MediumLevel}=4.29$) as compared to the "Successful Robot" group ($Mdn_{LowLevel}=4.33$, $Mdn_{MediumLevel}=4.33$). These differences were not significant in each of the separate sub measures (understanding and comfortability) and in the combined ease-of-use measure.

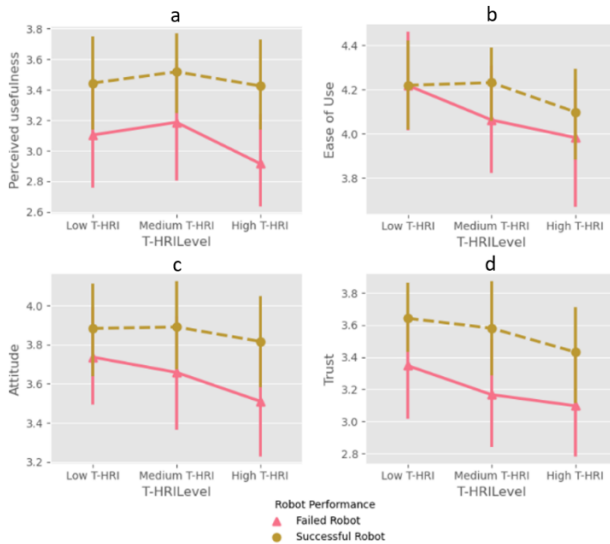

Figure 2 – Perceived usefulness (a), ease of use (b), attitude (c) and trust (d) by group and T-HRI level

### C. Attitude

Similar to the other measures the attitude measure was lower in the "Failed Robot" group than in the "Successful Robot" group for all three T-HRI levels (**Figure 2**). The median attitude in the low, medium, and high levels were 3.79, 3.73, 3.6, respectively, compared to the means in the "Successful Robot" that were 3.88, 3.96, 3.88. The attitude measure did not yield significant differences between the groups. However, analyses of the sub measures of the trust component resulted in more significant results (**Figure 2**). The Mann-Withney test showed differences between the "Failed Robot" and the "Successful Robot" group in the medium and high T-HRI levels ($p_{MediumLevel}=0.08$, $p_{HighLevel}=0.06$).

### D. Variance

There was a significant difference ($p_{Mann-Withney}=0.03$) between the standard deviation of the measures in the "Failed Robot" and the "Successful Robot" groups. The "Failed Robot" group's mean standard deviation (0.75) was higher than in the "Successful Robot" group (0.66) implying that a failing robot can influence differently different users.

## IV. CONCLUSIONS & DISCUSSION

There were differences between the "Failed Robot" group and the "Successful Robot" group in terms of perceived usefulness, ease-of-use, and attitude. Based on all measures, the "Failed Robot" group rated the robot lower than the "Successful Robot" group, regardless of T-HRI level. Failures as expected impact users perception of the robot.

Differences in the measures based on the T-HRI levels reveal a decreasing trend for both groups from the low to the high T-HRI level. This is particularly true for the "Failed Robot" group. For the ease-of-use measure, the low T-HRI level was ranked better than the high T-HRI level for both groups. At the low T-HRI level, the ease-of-use for both groups was similar. However, at the medium T-HRI level the "Failed Robot" group ranked it lower, and the "Successful Robot" group ranked it higher. At the high T-HRI level, the ease-of use decreased for both groups. The attitude measure has the same trends along the T-HRI levels and the participant groups. Meaning that when the robots failed the participants did not appreciate a higher T-HRI level. This might be due to the robot being perceived as untrustworthy, and a higher level of T-HRI leading to increased frustration. With the higher T-HRI level, the robot appears to be more intelligent, which leads to higher expectations, yet the robot continues to fail.

It should be considered that the higher T-HRI level is rated lower also in the "Successful Robot" group in several measures and that the differences between the T-HRI levels were not significant within the groups.

The T-HRI levels that were implemented and presented in this study should be improved to achieve more meaningful and significant results. This will especially be important for the application of the physical trainer towards it target population, older adults, who are less aware of technology. As preferences of LOT levels can be effected by users' age [11], the T-HRI levels must be adjusted more carefully to the task of a robotic trainer and for the users population (young adults or older adults).

A limitation in this study was that the failure was not controlled, therefore the results interpreted be taken under this assumption. The differences might have been clearer and more significant between the T-HRI levels and the participant groups if the failures would have been controlled.

The variance within the "Failed Robot" group was much higher as compared to the "Successful Robot" group indicating a more heterogenous group. Different users react differently to failures; this can result from various reasons such as familiarity with technology and robots, or personal traits. Therefore, studies should consider these effects and implement these considerations in the design of HRI.

## V. FUTURE STUDY

In future studies the robotic trainer and the T-HRI levels will be examined under controlled failures scenarios. Moreover, the implementation of the T-HRI levels should be improved to better enhance the difference between them. One option might be to implement only two T-HRI levels (low and high), to decrease the number of sessions for each participant hoping this will make the differences between the levels more meaningful and noticeable. This will probably be especially important once we start experimenting with older adults similar to [20]. Furthermore, future study should incorporate information about the failures of the robot as part of the T-HRI levels. The transparency model should include adjustments of robot's behaviors according to the interaction.

The future study will be focused on older adults as the target population of Gymmy. We expect all these trends to be amplified in such a group due to their lack of familiarity with technology. This will also enable to compare between different age groups which is important as noted in previous studies [10] and enable to examine users' preferences. Based on the initial results in this study, research should also examine how different personality traits affect the attitude toward robot's failure and to use these finding in the design of social assistive robots [12]. Moreover, the exercise background of the participants needs to be collected (how many times they train in a week, which kind of activities), to allow examination of this affect. Another interesting issue that raises from this research findings is how user's expectations can influence the interaction and attitude toward the robot and its failures. Overall, this initial examination of T-HRI levels along with the incorporation of robot failures has raised a number of interesting new directions for future research.